\documentclass[letterpaper, 10pt, conference]{ieeeconf}      % Use this line for a4 paper
\IEEEoverridecommandlockouts
\usepackage{cite}
\usepackage[dvips]{graphicx}
\usepackage{multirow}
\usepackage[top=0.75in, left=0.75in, right=0.75in, bottom=0.75in]{geometry}
\usepackage{times}
\usepackage{makeidx}  % allows for indexgeneration
\usepackage{amsmath}
\usepackage{amssymb}
\usepackage{latexsym}
\usepackage{epsfig}
\usepackage{verbatim}
\usepackage{psfrag}
\usepackage{graphicx}
\usepackage{subfigure}
\usepackage{cite}
\usepackage{algorithmic}
\usepackage{array}
\usepackage{bm}
\usepackage{xcolor}
\usepackage{arydshln}
\title{\LARGE \bf Trajectory Generation for Underactuated Multirotor Vehicles with Tilted Propellers via a Flatness-based Method}

\author{Bingguo Mu and Pakpong Chirarattananon% <-this % stops a space
\thanks{This work was substantially supported by the Research Grants Council of the Hong Kong Special Administrative Region of China (grant number CityU-11274016).}% <-this % stops a space
\thanks{The authors are with the Department of Biomedical Engineering, City University of Hong Kong, Hong Kong SAR, China (emails: bingguomu2-c@my.cityu.edu.hk and  pakpong.c@cityu.edu.hk).}
}

\begin{document}

\maketitle
\thispagestyle{empty}
\pagestyle{empty}

\begin{abstract}
This paper considers a class of rotary-wing aerial robots with unaligned propellers. By studying the dynamics of these vehicles, we show that the position and heading angle remain flat outputs of the system (similar to conventional quadrotors). The implication is that they can be commanded to follow desired trajectory setpoints in 3D space. We propose a numerical strategy based on the collocation method to facilitate the trajectory generation. This enables convenient computation of the nominal robot's attitude and control inputs. The proposed methods are numerically verified for three multirotor robots with different dynamics. These include a tricopter with a tilting propeller, and a quadrotor and a hexacopter with unparalleled propellers.
\end{abstract}

\section{Introduction}

In the past decade, we have witnessed an unrivaled popularity of multirotor vehicles. The widespread adoption is attributed by the capability to vertically takeoff and land, as well as its relative ease of use. The mechanical simplicity of conventional quadrotors captures the attention of researchers and engineers and facilitates them to further develop the platforms for numerous applications \cite{hsiao2018ceiling, chirarattananon2018direct, morrell2018differential, papachristos2014efficient}.

More recently, several researchers have pioneered multirotor robots with alternative designs. Among these, the ability to perform thrust-vectoring is becoming prevalent. Additional actuators enable these robots to tilt the propellers, resulting in holonomic aerial platforms \cite{papachristos2014efficient,ramp2015modeling,ryll2015novel,brescianini2018computationally,rajappa2015modeling}. While this ability of thrust-vectoring inevitably comes with added weight and energy consumption, it allows robots to interact more proficiently with the environment \cite{brescianini2018computationally}. For instance, the robot forcefully interacts with an object via a manipulator \cite{papachristos2014efficient}.

Unlike fully-actuated robots with thrust vectoring capabilities, underactuated aerial vehicles are unable to track arbitrary trajectories in $SE(3)$. To date, there has been little development when it comes to aerial vehicles with tilted propellers that are not fully actuated \cite{efraim2015quadrotor, ansari2017adaptive}. In such vehicles, the direction of the total thrust varies in the body frame and the rotational and translational dynamics are sophisticatedly coupled. Without consideration of the distinct dynamic model and suitable control strategies, the progress in flight of multirotor vehicles with unaligned propellers remains infancy.

\begin{figure}
\centering
\psfrag{sys}[c][c][0.9]{body frame}
\psfrag{world}[c][c][0.9]{inertial frame}
\psfrag{zw}[c][c][0.9]{$Z_\text{W}$}
\psfrag{yw}[c][c][0.9]{$Y_\text{W}$}
\psfrag{xw}[c][c][0.9]{$X_\text{W}$}
\psfrag{zb}[c][c][0.9]{$Z_\text{B}$}
\psfrag{yb}[c][c][0.9]{$Y_\text{B}$~~~~~~~~~~~~~~~~~~~~~~~~~~~~~~~}
\psfrag{xb}[c][c][0.9]{~~~~~~~~~~~~~~$X_\text{B}$}
\psfrag{T1}[c][c][0.9]{$T_1$}
\psfrag{T2}[c][c][0.9]{$T_2$}
\psfrag{T3}[c][c][0.9]{$T_3$}
\psfrag{T4}[c][c][0.9]{$T_4$}
\psfrag{T}[c][c][0.9]{$T$}
\psfrag{mg}[c][c][0.9]{$mg$}
\includegraphics[width=70mm]{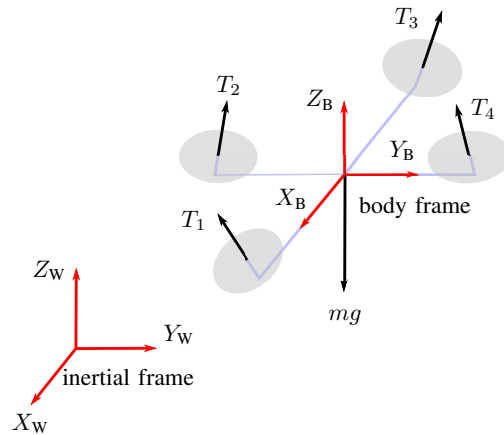}
\caption{The tilted thrusts from the tilted propellers.}\label{fig.thrusts}\vspace{-4mm}
\end{figure}

In this paper, we investigate the flight dynamics of aerial vehicles with unaligned thrusts as depicted in Fig. \ref{fig.thrusts}. For regular underacuated robots, it has been shown that the position and heading angle form a set of flat outputs of the system \cite{mellinger2012trajectory,morrell2018differential}. This means the aerial robots can be controlled to follow 3D trajectory setpoints, rendering them highly versatile platforms for various applications.

Motivated by the benefits of flatness property, we attempt to show that the position and the heading angle remain the flat outputs of the underacuated robots with tilted thrusts. By doing so, corresponding attitude states and control inputs can be found. This trajectory generation strategy, when combined with a compatible flight controller, would potentially lead to highly versatile robots that can perform mid-flight reconfiguration without sacrificing on the controllability.

In the next section, we first outline the dynamic model of robots with unaligned thrusts. We highlight the differences between our model and a model of a regular quadrotor. In section \ref{sec.trajectory_gen}, we show that the position and heading angle are flat outputs of the system. Then, a computational method for generating the desired trajectory is proposed. The framework is verified using numerical simulations with three multirotor vehicles with different dynamics.

\section{Dynamic Model}

To describe the flight dynamics of a multirotor vehicle with $N$ propellers, we define two coordinate frames: the world frame $\{X_\text{W},Y_\text{W},Z_\text{W}\}$ and the body-fixed frame $\{X_\text{B},Y_\text{B},Z_\text{B}\}$ located on the center of mass (CM) of the robot. A rotation matrix $\bm{R}:=[X_{\text{B}},Y_{\text{B}},Z_{\text{B}}]\in SO(3)$ is defined to relate the attitude of the body frame with respect to the world frame. Let $\bm{\omega}=[\bm{\omega}_x,\bm{\omega}_y,\bm{\omega}_z]$ represents the angular velocity of the robot and $\hat{\bm{\omega}}$ be its associated skew-symmetric matrix, then the time derivative of $\bm{R}$ satisfies
\begin{equation}
\dot{\bm{R}} = \bm{R}\hat{\bm{\omega}}.\label{eqn.dotomega}
\end{equation}
For simplicity, the rotation matrix $\bm{R}$ is represented by three Euler angles, yaw ($\psi$),  pitch ($\theta$) and roll ($\phi$) in the z-y-x sequence with $\bm{\theta}:=[\phi,\theta,\psi]^\text{T}$, then we have $\bm{R}(\bm{\theta})=\bm{R}_{z}(\psi)\bm{R}_{y}(\theta)\bm{R}_{x}(\phi)$. The relationship between these three angles and the angular velocity $\bm{\omega}$ is
\begin{equation}
\bm{\omega}
=
\left[
  \begin{array}{ccc}
    1 & 0 & -\sin(\theta) \\
    0 & \cos(\phi) & \sin(\phi)\cos(\theta) \\
    0 & -\sin(\phi) & \cos(\phi)\cos(\theta) \\
  \end{array}
\right]
\dot{\bm{\theta}}
:=\bm{E}(\bm{\theta})\dot{\bm{\theta}}.\label{eqn.euler_angles}
\end{equation}
Let $\bm{P}=[x,y,z]^\text{T}$ denote the position of the vehicle in the world frame. The robot's acceleration depends on the vector sum of all propeller's thrusts. Let $T_i$ denote the magnitude of the thrust from the $i^\text{th}$ propeller and $\bm{v}_i$ be a unit vector describing the direction of $T_i$ in the body frame. The translational dynamics of the robot with mass $m$ are
\begin{equation}
\begin{split}
m \ddot{\bm{P}}-m \bm{g}
&=\bm{R}\sum^{N}_{i=1}(\bm{v}_iT_i)\\
&= \bm{R}\underbrace{\left [\bm{v}_1,\bm{v}_2,\cdots,\bm{v}_N \right ]}_{\bm{A}_{3\times N}} \underbrace{\left[T_{1},T_{2},\cdots,T_{N}\right]^\text{T}}_{\bm{u}},
\end{split}\label{eqn.translation_dynamics}
\end{equation}
where we define the matrix $\bm{A}$ to describe the alignment of the propellers and $\bm{g}=[0,0,-g]^\text{T}$ is the gravity vector.

To obtain the rotational dynamics, we let $\bm{r}_i$ denote the moment arm of the $i^\text{th}$ propeller in the body frame. To capture the torque contributed by each propeller, we define the matrix $\bm{M}_i$ for the $i^\text{th}$ propeller as
\begin{equation}
\bm{M}_i=
\hat{\bm{r}}_i+c_i\bm{\mathcal{I}}_{3\times 3},\label{eqn.Mmatrix}
\end{equation}
where $\bm{\mathcal{I}}$ is an identity matrix, $\hat{\bm{r}}_i$ is the skew-symmetric matrix of $\bm{r}_i$, and $c_i$ is the aerodynamic constant denoting the ratio of the drag torque to the thrust from the $i^\text{th}$ propeller. The sign of $c_i$ is determined by the spinning direction of the propeller. Thus, the rotational dynamics of the robot follow
\begin{equation}
\begin{split}
\sum^N_{i=1}(\bm{r}_i\times(\bm{v}_iT_i)+c_i\bm{v}_iT_i)
&=\underbrace{[\bm{M}_1\bm{v}_1,\cdots,\bm{M}_N\bm{v}_N]}_{\bm{B}_{3\times N}}\bm{u}\\
&=\bm{J}\dot{\bm{\omega}}+\hat{\bm{\omega}}\bm{J}\bm{\omega}.\label{eqn.orientation_dynamics}
\end{split}
\end{equation}
The translational and rotational dynamics from equations \eqref{eqn.translation_dynamics} and \eqref{eqn.orientation_dynamics} can be consolidated as
\begin{equation}
\underbrace{
\left[
  \begin{array}{cc}
    \bm{A} & \bm{0} \\
    -\bm{B} & \bm{J} \\
  \end{array}
\right]}_{\bm{C}}
\left[
  \begin{array}{c}
    \bm{u} \\
    \dot{\bm{\omega}} \\
  \end{array}
\right]
=
\left[
  \begin{array}{c}
    m\bm{R}^\text{T}\left (\ddot{\bm{P}}-\bm{g}\right ) \\
    -\hat{\bm{\omega}}\bm{J}\bm{\omega} \\
  \end{array}
\right]. \label{eqn.dynamics}
\end{equation}
In the equation above, the orientations of the thrust vectors ($\bm{v}_i$'s) are embedded into matrices $\bm{A}$ and $\bm{B}$. One critical condition for the controllability is $\text{rank}(\bm{B})= 3$, which means the robot is capable of independently generating torques about its three body axes. For a typical multirotor robot, all propellers are aligned ($\bm{v}_1=\bm{v}_2=\cdots=\bm{v}_N$), and the rank of $\bm{A}$ is one. In such case, equation \eqref{eqn.translation_dynamics} reduces to  $\|m\ddot{\bm{P}}-m\bm{g}\|=\sum^N_{i=1}T_i$ and the orientation of the robot ($\bm{R}$) is uniquely and immediately determined from $\ddot{\bm{P}}-\bm{g}$ and the desired heading \cite{zhou2014vector}. In this work, we relax on the rank condition of $\bm{A}$ and investigate a class of multirotor robots of which all propellers are not necessarily aligned, or $\text{rank}(\bm{A})\neq 1$.

\section{Flatness-based Trajectory Generation} \label{sec.trajectory_gen}

In \cite{mellinger2012trajectory}, it has been shown that the dynamics of a regular quadrotor with four inputs are differentially flat with the position and heading of the robot constituting four flat outputs. The implication of the result is that, there always exists a feasible trajectory and admissible controls for the chosen flat outputs. With a compatible controller, the robot can be prescribed to follow a setpoint trajectory in 3D space. Without the flatness property, the task of commanding the underactuated robot to traverse from one setpoint to another setpoint requires one to solve the trajectory optimization problem and verify the feasibility. In this section, we employ the differential flatness framework to (i) show that for a class of multirotor robots with unaligned thrusts, there exist a feasible trajectory and control inputs for four flat outputs: position and heading, similar to a standard quadrotor; and (ii) propose a universal trajectory generation method for all related vehicles regardless of the rank condition of $\bm{A}$. 

\subsection{Differential Flatness}
For a dynamical system $\dot{\bm{x}}=f\left(\bm{x},\bm{u}\right)$, $\bm{x}\in \mathbb{R}^{m}$, $\bm{u}\in \mathbb{R}^{n}$, it is said to be differentially flat if there exists a flat output: $\bm{\sigma}=h(\bm{x},\dot{\bm{u}},\ddot{\bm{u}},\ldots,\bm{u}^{(p)})$, such that there exist functions $\Phi_{\bm{x}}$ and $\Phi_{\bm{u}}$ \cite{zhou2014vector,fliess1995flatness}:
\begin{equation}
\begin{split}
\bm{x}	&=\Phi_{\bm{x}}(\bm{\sigma},\dot{\bm{\sigma}},\ddot{\bm{\sigma}},\ldots,
\bm{\sigma}^{(q)}),\\
\bm{u}	&=\Phi_{\bm{u}}(\bm{\sigma},\dot{\bm{\sigma}},\ddot{\bm{\sigma}},\ldots,\bm{\sigma}^{(q)}),
\end{split}\label{eqn.differentially_flat}
\end{equation}
where $p$ and $q$ are finite numbers. Equation \eqref{eqn.differentially_flat} implies that if the flat output $\bm{\sigma}$ is found, the trajectories of the states and the control inputs are immediately deduced. This is beneficial for motion planning and stabilization of setpoint trajectories. The flatness property is also perceived as an extension of the controllability of linear systems.

In \cite{fliess1995flatness}, it has been shown that the dynamics of a regular quadrotor is 0-flat, with the corresponding flat output $\bm{\sigma}(t)$,
\iffalse
Generating an optimal trajectory for the flying vehicles usually becomes an optimization problem based on the performance requirements and input constraints \cite{mellinger2012trajectory,zhou2014vector,brescianini2018computationally}. The flight trajectories can be planned in the output space based on the differential flatness theory. We investigate the trajectory generation for a class of multirotor vehicles without considering the matrix $\bm{A}$ is full rank or not. In the generated trajectory, all states and control inputs of the multirotor vehicles can not be analytically expressed though, but they can be uniquely solved via applying the collocation method \cite{kelly2017introduction}.
\fi
\begin{equation}
\bm{\sigma}(t) = [x(t),y(t),z(t),\psi(t)]^\text{T}.\label{eqn.flat_output}
\end{equation}
As a result, a conventional quadrotor with paralleled thrusts is able to follow a reference trajectory. Without the flatness property, it is necessary to employ trajectory optimization methods to search and verify that there exists a feasible trajectory to command the robot from the starting setpoint to the desired setpoint.

When propellers on a robot are no longer aligned, the translational dynamics are tightly coupled with the rotational dynamics. For instance, equations \eqref{eqn.translation_dynamics} and \eqref{eqn.orientation_dynamics} imply that it is impossible for the robot to retain a net zero acceleration in the body frame when it generates torque. The difference from a conventional quadrotor means that the flatness proof in \cite{mellinger2012trajectory} is no longer valid. More specifically, the difference in the dynamics equations belonging to the two scenarios is captured into $\bm{A}$ and $\bm{B}$, especially in the rank of matrix $\bm{A}$.

\subsection{Existence of Flat Outputs} \label{sec.existence}
Herein, we revisit the dynamics of multirotor robots with unaligned thrusts and show that the previously identified output in equation \eqref{eqn.flat_output} is still the flat output to the revised dynamics. Without loss of generality, the analysis in this section assumes $N=4$. In cases where $N>4$, the method can be conveniently adapted as presented in some examples implementation later in Section \ref{sec:verification}. For illustrative purposes, we formally define a state vector $\bm{x}=[\bm{P},\dot{\bm{P}},\bm{\theta},\bm{\omega}]^\text{T}$. Next, we consider two scenarios: $\text{rank}(\bm{A})=3$ and $\text{rank}(\bm{A})=2$. 

\iffalse
To begin, we formally , where Euler angles $\bm{\theta}=\left[\theta_{x},\theta_{y},\theta_{z}\right]$ are used to describe the attitude of the robot such that $\bm{R}=\bm{R}\left(\theta_{x},\theta_{y},\theta_{z}\right)$. Other representations can be used without loss of generality. It is evident that the flat output defined by equation \eqref{eqn.flat_output} is fully determined from the state vector.

The translational states including position $\bm{P}$, and velocity $\dot{\bm{P}}$ of the vehicles are $\bm{\sigma}_{1:3}$, $\dot{\bm{\sigma}}_{1:3}$ and $\ddot{\bm{\sigma}}_{1:3}$.

We determine the rotational states and control input of the multirotor robots by considering the rank of matrix $\bm{A}$ in three cases. Note, matrix $\bm{B}$ is always assumed to full rank.
\fi

\subsubsection{$\text{rank}(\bm{A})=3$}

In this circumstance, the rank of the $6\times (N+3)$ matrix $\bm{C}$ in equation \eqref{eqn.dynamics} is six. With the yaw angle of the robot as part of the flat output $\psi(t)=\bm{\sigma}_4(t)$, this results in one further constraint on $\bm{\omega}$, and equivalently on $\dot{\bm{\omega}}$, through equation \eqref{eqn.euler_angles} by the fact that $\ddot{\psi}=\ddot{\bm{\sigma}}_4$. Treating the $7\times 1$ vector $[\bm{u},\dot{\bm{\omega}}]^\text{T}$ in equation \eqref{eqn.dynamics} as an unknown, the flat output provides balanced constraints to directly evaluate $\bm{u}$ and $\bm{\omega}$ in the form
\begin{equation}
[\bm{u},\dot{\bm{\omega}}]^\text{T} = f(\bm{\sigma},\bm{\theta},\bm{\omega}), \label{eqn.rank3_solution}
\end{equation}
such that there always exist $\bm{u}$ and $\dot{\bm{\omega}}$ given $\bm{\sigma}$, $\bm{\theta}$, and $\bm{\omega}$. It is then straightforward to solve the above equation as a second-order differential equation to simultaneously obtain the states and control inputs given suitable initial conditions. In other words, there exists a numerical function $\Phi_{\bm{x},\bm{u}}$
\begin{equation}
[\bm{x},\bm{u}]^\text{T}=\Phi_{\bm{x},\bm{u}}(\bm{\sigma},\dot{\bm{\sigma}},\ddot{\bm{\sigma}},\ldots,\bm{\sigma}^{(q)}), \label{eqn.output_fn_rank3}
\end{equation}
similar to equation \eqref{eqn.differentially_flat}. It can be concluded the system with $\text{rank}(\bm{A})=3$ is differentially flat with the proposed output.

\subsubsection{$\text{rank}(\bm{A})=2$}

In this case, the rank of $\bm{C}$ is 5 and the above method to simultaneously solve for $\bm{u}$ and $\dot{\bm{\omega}}$ no longer applies. Instead, we consider the singular-value decomposition: $\bm{A}=\bm{Q}\bm{\Sigma}\bm{V}^\text{T}$, where $\bm{Q}$ and $\bm{V}$ are $3\times 3$ and $N\times N$ orthogonal matrices, and $\bm{\Sigma}$ is a $3\times N$ matrix with two non-zero elements ($\bm{\Sigma}_{1,1}=\lambda_1$ and $\bm{\Sigma}_{2,2}=\lambda_2$). Exploiting the fact that $\bm{Q}\in SO(3)$, we redefine the body frame with the rotation matrix $\bar{\bm{R}}(\bm{\theta})=\bm{R}(\bm{\theta})\bm{Q}$ and the input vector $\bar{\bm{u}}=\bm{V}^\text{T}\bm{u}=\left[\bar{T}_{1},\bar{T}_{2},\cdots,\bar{T}_{N}\right]^\text{T}$. The dynamics of the vehicle becomes
\begin{align}
m \bar{\bm{R}}^\text{T} (\ddot{\bm{\sigma}}_{1:3}-\bm{g}) &= \bm{\Sigma}\bar{\bm{u}} \label{eqn.redefined_t_dynamics}, \\
\bar{\bm{J}}\dot{\bar{\bm{\omega}}}+\hat{\bar{\bm{\omega}}}\bar{\bm{J}}\bar{\bm{\omega}}&=\bar{\bm{B}}\bar{\bm{u}}, \label{eqn.redefined_att_dynamics}
\end{align}
where $\bar{\bm{J}}=\bm{Q}^\text{T}\bm{J}\bm{Q}$, $\bar{\bm{B}}=\bm{Q}^\text{T}\bm{B}\bm{V}$, and $\bar{\bm{\omega}}=\bm{Q}^\text{T}\bm{\omega}$ are re-defined variables in the new body frame. Equation \eqref{eqn.redefined_t_dynamics} can be manipulated further as
\begin{equation}
{\bm{Q}}^\text{T}\bm{R}_{x}^\text{T}(\phi)\bm{R}_{y}^\text{T}(\theta)\bm{R}_{z}^\text{T}(\bm{\sigma}_4)\frac{\ddot{\bm{\sigma}}_{1:3}-\bm{g}} {\|(\ddot{\bm{\sigma}}_{1:3}-\bm{g})\|}=\frac{[\lambda_{1}\bar{T}_{1},\lambda_{2}\bar{T}_{2},0]^\text{T}}{{m\|(\ddot{\bm{\sigma}}_{1:3}-\bm{g})\|}}. \label{eqn.rank2_derivation01}
\end{equation}
Leveraging the orthorgonality of rotation matrices, we define $\bm{f}(\bm{\sigma})=\bm{R}_{z}^\text{T}(\bm{\sigma}_4)(\ddot{\bm{\sigma}}_{1:3}-\bm{g})$ and $\Theta_T$ that satisfies $\cos(\Theta_T)=\lambda_1\bar{T}_1/\|m\bm{f}\|$, $\sin(\Theta_T)=-\lambda_2\bar{T}_2/\|m\bm{f}\|$. It follows that
\begin{equation}
\begin{split}
{\bm{Q}}^\text{T}\bm{R}_{x}^\text{T}(\phi)\bm{R}_{y}^\text{T}(\theta){\bm{f}}/{\|\bm{f}\|}&={[\cos(\Theta_T),-\sin(\Theta_T),0]^\text{T}}\\
&=\bm{R}^\text{T}_{z}(\Theta_T)[1,0,0]^\text{T}. \label{eqn.rank2_derivation02} 
\end{split}
\end{equation}
With $\bar{T}_i$'s not yet determined (and likewise $\Theta_T$) equation \eqref{eqn.rank2_derivation02} imposes one DoF constraint on the rotation matrix $\bm{R}(\bm{\theta})$. Next, we consider an alternative parametrization of $\bm{R}(\bm{\theta})$. We propose $\bm{R}(\bm{\Theta}):=\bm{R}_z(\psi)\bm{R}(\Theta_f)\bm{R}_z(\Theta_T)=\bar{\bm{R}}(\bm{\theta})$ for $\bm{\Theta}:=[\psi,\Theta_f,\Theta_T]$. As a consequence, equation (\ref{eqn.rank2_derivation02}) becomes
\begin{equation}
\bm{R}_z^\text{T}(\Theta_T)\bm{R}^\text{T}(\Theta_f) {\bm{f}}/{\|\bm{f}\|}=\bm{R}^\text{T}_{z}(\Theta_T)[1,0,0]^\text{T}. \label{eqn.rank2_derivation03} 
\end{equation}
From the equation above, it is straightforward to obtain $\bm{R}(\Theta_f)$ by solving the equation 
\begin{equation}
\bm{R}^\text{T}(\Theta_f){\bm{f}}/{\|\bm{f}\|}=[1,0,0]^\text{T}\label{eqn.rank2_derivation08} .
\end{equation}
Notice that $\Theta_f$ depends entirely on $\bm{f}(\bm{\sigma})$, that is $\Theta_f=\Theta_f(\bm{\sigma})$. This $\Theta_f$ captures the one DoF information on the robot's orientation embedded in \eqref{eqn.rank2_derivation02}. This leaves only one DoF ($\Theta_T$) undetermined.

To incorporate this result with the attitude dynamics from equation \eqref{eqn.redefined_att_dynamics}, we analyze the time derivative of $\bm{R}(\bm{\Theta})$. Similar to the process of obtaining equation \eqref{eqn.euler_angles}, we find
\begin{equation}
\begin{split}
\bar{\bm{\omega}}
=&\bm{R}_{z}^\text{T}(\Theta_T)\bm{R}^\text{T}(\Theta_f)\bm{R}_{z}^\text{T}(\bm{\sigma}_4)[0,0,\dot{\bm{\sigma}}_4]^\text{T}\\
&+\bm{R}_{z}^\text{T}(\Theta_T)(\bm{R}^\text{T}(\Theta_f)\dot{\bm{R}}(\Theta_f))^\vee\\
&+\bm{R}_{z}^\text{T}(\Theta_{T})[0,0,\dot{\Theta}_{T}]^\text{T} \\
=&[0,0,\dot{\Theta}_{T}]^\text{T}+\bm{R}^\text{T}_{z}(\Theta_T)\bm{R}^\text{T}(\Theta_f)[0,0,\dot{\bm{\sigma}}_4]^\text{T}\\
&+\bm{R}_{z}^\text{T}(\Theta_T)(\bm{R}^\text{T}(\Theta_f)\dot{\bm{R}}(\Theta_f))^\vee \\
=&[0,0,\dot{\Theta}_{T}]^\text{T}+\bm{\xi}(\dot{\bm{\sigma}}_4,\Theta_f,\dot{\Theta}_f,\Theta_T),
\end{split}\label{eqn.rank2_derivation04}
\end{equation}
where the operation $(\cdot)^\vee$ is the inverse of the hat operation:  $(\hat{\cdot})^\vee=(\cdot)$, and $\bm{\xi}$ is a $3\times 1$ vector from the simplification. The time derivative of $\bar{\bm{\omega}}$ can be computed as
\begin{equation}
\dot{\bar{\bm{\omega}}}= [0,0,\ddot{\Theta}_{T}]^\text{T}+\dot{\bm{\xi}}(\dot{\bm{\sigma}}_4,\ddot{\bm{\sigma}}_4,\Theta_f,\dot{\Theta}_f,\ddot{\Theta}_f,\Theta_T,\dot{\Theta}_T).\label{eqn.rank2_derivation05}
\end{equation}
Finally, with the new parametrization of $\bar{\bm{\omega}}(\bm{\sigma},\bm{\Theta},\dot{\bm{\Theta}})$, the attitude dynamics from equation \eqref{eqn.redefined_att_dynamics} becomes
\begin{equation}
\bar{\bm{B}} 
\left[\begin{array}{c}
\cos\Theta_{T}\frac{\|m\bm{f}\|}{\lambda_{1}}\\
-\sin\Theta_{T}\frac{\|m\bm{f}\|}{\lambda_{2}}\\
\bar{T}_{3}\\
\bar{T}_{4}\\
\end{array}
\right]
-\hat{\bar{\bm{\omega}}}\bar{\bm{J}}\bar{\bm{\omega}}-\bar{\bm{J}}\dot{\bm{\xi}}=\bar{\bm{J}}
\left[\begin{array}{c}
0\\
0\\
\ddot{\Theta}_{T}\\
\end{array}\right]. \label{eqn.rank2_derivation06}
\end{equation}
We pre-multiply equation \eqref{eqn.rank2_derivation06} with $\bar{\bm{J}}^{-1}$ and use the top two rows to solve for $\bar{T}_3$ and $\bar{T}_4$ in terms of $\bm{\sigma}, \Theta_T, \dot{\Theta}_T$. The results are substituted back into the last row, which can eventually be expressed as
\begin{equation}
\ddot{\Theta}_T = f(\bm{\sigma}, \Theta_T, \dot{\Theta}_T)\label{eqn.rank2_derivation07} .
\end{equation}
With suitable initial conditions, $\bm{\sigma}_4$, and previously determined $\Theta_f$, solving the respective second-order differential equation provides the complete attitude trajectories. The input $\bm{\bar{u}}$ can then be determined from $\Theta_T$. That is, the control input $\bm{u}$ can be found from the flat outputs. This proves that the system is differentially flat with respect to $\bm{\sigma}$.

\subsection{Trajectory Generation with Direct Collocation}

The results in the previous section verify that dynamics described by equation \eqref{eqn.dynamics} are differentially flat with the proposed flat outputs in equation \eqref{eqn.flat_output} regardless of the rank of $\bm{A}$. Nevertheless, different approaches are required to numerically compute the corresponding states and inputs.

To facilitate the process, we propose to unify the trajectory generation approach for all cases (including the case for a regular quadrotor or $\text{rank}(\bm{A})=1$) by employing the direct collocation method for the trajectory generation.

The application of the direct collocation, traditionally used for trajectory optimization, hinges on the dynamics in equation \eqref{eqn.dynamics}. While this equation cannot be solved analytically when $\text{rank}(\bm{A})\neq3$, it can be solved numerically for the entire trajectory by dividing the trajectory into multiple collocation points. For demonstrative purposes, we show an example implementation of the direct transcription using Euler integration. This begins by dividing the trajectory into $n$ time steps: $t=t_i$ for $i=1,2,\ldots,n$. In the discrete-time domain of sample time $t_s$, equation \eqref{eqn.dynamics} is approximated as
\begin{equation}
\begin{split}
\bm{A}\bm{u}_i  &= m\bm{R}(\bm{\theta}_i)^\text{T}(\ddot{\bm{\sigma}}_{1:3,i}-\bm{g}),\\
\bm{J}(\bm{\omega}_{i+1}-\bm{\omega}_{i}) &= t_s(\bm{B}\bm{u}_i -\hat{\bm{\omega}}_i\bm{J}\bm{\omega}_i),
 \label{eqn.dynamics_direct_transcription}
\end{split}
\end{equation}
which together count as $6n$ scalar equations. The complete dynamics also include additional $3n$ equations from
\begin{equation}
\bm{\theta}_{i+1}-\bm{\theta}_{i} = t_s\bm{E}^{-1}(\bm{\theta}_i)\bm{\omega}_i,\label{eqn.angle_direct_transcription}
\end{equation}
where the relationship between $\bm{\theta}_i$ and $\bm{\omega}_i$, $\bm{E}(\bm{\theta}_i)$, is determined by equation \eqref{eqn.euler_angles}. The yaw angle setpoint implies
\begin{equation}
\psi_i=\bm{\sigma}_{4,i}.\label{eqn.yaw_direct_transcription}
\end{equation}
If we treat $\bm{u}_i$, $\bm{\theta}_i$, and $\bm{\omega}_i$ as unknown variables, equations \eqref{eqn.dynamics_direct_transcription}-\eqref{eqn.yaw_direct_transcription} consist of $10n+6$ unknowns and $10n$ constraints. Additional six initial conditions can be chosen and the nonlinear program is well defined. Owing to the proofs in Section \ref{sec.existence}, the existence of the solution is guaranteed.

In practice, a more accurate implementation, e.g., Hermite-Simpson collocation, is preferred. Fortunately, there are a variety of open-source and commercial software packages capable of solving such nonlinear programs. These solvers are conventionally designed for trajectory optimization tasks. This involves an optimization of a cost function for a problem with more variables than constraints, in which oftentimes the feasible solution may not exist. Hence, it is important to emphasize that the implementation of the collocation method for our trajectory generation is different, as our nonlinear problem is not underdetermined and a feasible solution is guaranteed to exist. The benefit of using the collocation method is predominantly the ease of implementation.
\begin{figure}[t]
\centering
\psfrag{ts}[c][c][0.9]{t (s)}
\psfrag{s0g}[c][c][0.9]{$\bm{\sigma}$}
\psfrag{s1g}[c][c][0.9]{$\dot{\bm{\sigma}}$}
\psfrag{s2g}[c][c][0.9]{$\ddot{\bm{\sigma}}$}
\psfrag{s3g}[c][c][0.9]{$\bm{\sigma}^{(3)}$}
\psfrag{s4g}[c][c][0.9]{$\bm{\sigma}^{(4)}$}
\psfrag{sg1}[c][c][0.9]{$\bm{\sigma}_1$}
\psfrag{sg2}[c][c][0.9]{$\bm{\sigma}_2$}
\psfrag{sg3}[c][c][0.9]{$\bm{\sigma}_3$}
\psfrag{sg4}[c][c][0.9]{$\bm{\sigma}_4$}
\includegraphics[width=85mm]{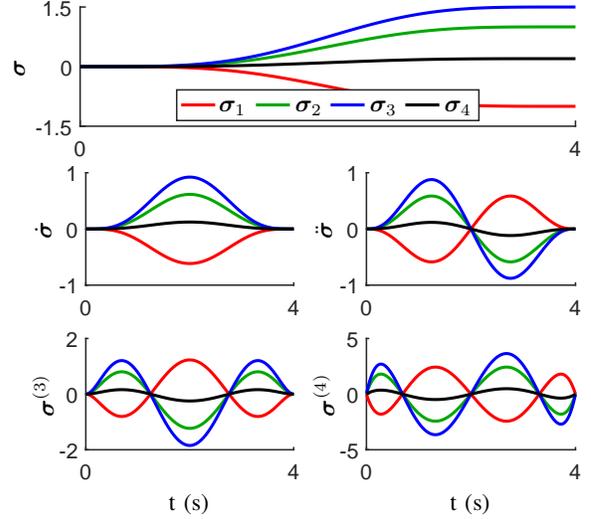}
\caption{A feasible trajectory $\bm{\sigma}$ (using standard units, meter for $\bm{\sigma}_{1:3}$ and radian for $\bm{\sigma}_4$) and its time-derivatives $\dot{\bm{\sigma}}$, $\ddot{\bm{\sigma}}$, $\bm{\sigma}^{(3)}$, $\bm{\sigma}^{(4)}$.}\label{fig.trajectory}\vspace{-2mm}
\end{figure}
\begin{figure*}[t]
\centering
\psfrag{bar}[c][c][0.9]{10 cm}
\psfrag{ap}[c][c][0.9]{$\alpha$}
\psfrag{Ta}[c][c][0.9]{$\bm{v}_\alpha$}
\psfrag{agt1}[c][c][0.6]{$\begin{bmatrix}0 \\ 0 \\ 1 \end{bmatrix}$}
\psfrag{agt2}[c][c][0.6]{$\begin{bmatrix}0 \\ 0 \\ 1 \end{bmatrix}$}
\psfrag{agt3}[c][c][0.6]{$\begin{bmatrix}0 \\ 0 \\ 1 \end{bmatrix}$~~~~~~~~~~~~~~~~~~~~~~~~~~~~~~~~~~~~~~~~}
\psfrag{agt4}[c][c][0.6]{$[0,-1,0]^\text{T}$}
\psfrag{agt5}[c][c][0.6]{$\begin{bmatrix}0 \\ -\sin(\alpha) \\ \cos(\alpha) \end{bmatrix}$}
\psfrag{agq1}[c][c][0.6]{$\begin{bmatrix}0.20 \\ 0 \\ 0.98 \end{bmatrix}$}
\psfrag{agq2}[c][c][0.6]{$\begin{bmatrix}0 \\ 0.30 \\ 0.96 \end{bmatrix}$}
\psfrag{agq3}[c][c][0.6]{$\begin{bmatrix}0.30 \\ 0 \\ 0.96 \end{bmatrix}$}
\psfrag{agq4}[c][c][0.6]{$\begin{bmatrix}0 \\ -0.10 \\ 0.99 \end{bmatrix}$}
\psfrag{agh1}[c][c][0.6]{$\begin{bmatrix}0.20 \\ 0 \\ 0.98 \end{bmatrix}$}
\psfrag{agh2}[c][c][0.6]{$\begin{bmatrix}0 \\ 0.30 \\ 0.96 \end{bmatrix}$}
\psfrag{agh3}[c][c][0.6]{$\begin{bmatrix}0 \\ -0.10 \\ 0.99 \end{bmatrix}$}
\psfrag{agh4}[c][c][0.6]{$\begin{bmatrix}0.20 \\ 0 \\ 0.98 \end{bmatrix}$}
\psfrag{agh5}[c][c][0.6]{$\begin{bmatrix}0 \\ 0.10 \\ 0.99 \end{bmatrix}$}
\psfrag{agh6}[c][c][0.6]{$\begin{bmatrix}0 \\ 0.20 \\ 0.98 \end{bmatrix}$}
\psfrag{rq1}[c][c][0.6]{$[19,0,0]^\text{T}$~~~~~~~~~~~~~~~~~~~~~~~~~}
\psfrag{rq2}[c][c][0.6]{$[0,-19,0]^\text{T}$}
\psfrag{rq3}[c][c][0.6]{~~~~~~~~~~~$[-19,0,0]^\text{T}$}
\psfrag{rq4}[c][c][0.6]{$[0,19,0]^\text{T}$}
\psfrag{rt1}[c][c][0.6]{$[-10,-17,0]^\text{T}$}
\psfrag{rt2}[c][c][0.6]{$[-10,17,0]^\text{T}$~~~~~~~}
\psfrag{rt3}[c][c][0.6]{$[19,0,0]^\text{T}$~~~~~~~~~~~~~~~~~~~~~~~~~~~}
\psfrag{rh1}[c][c][0.6]{~~~~~~~~~~~~~~~$[-19,0,0]^\text{T}$}
\psfrag{rh2}[c][c][0.6]{$[-9,17,0]^\text{T}$}
\psfrag{rh3}[c][c][0.6]{$[9,17,0]^\text{T}$}
\psfrag{rh4}[c][c][0.6]{$[19,0,0]^\text{T}$~~~~~~~~~~}
\psfrag{rh5}[c][c][0.6]{~~~~~~~~~~~~~~~~~~~~~~~~~~~~~~~~~~~~~$[9,-17,0]^\text{T}$}
\psfrag{rh6}[c][c][0.6]{~~~~~~~~~~~~~~~~~~~~~~~~~$[-9,-17,0]^\text{T}$}
\psfrag{agt}[c][c][0.6]{120$^\circ$}
\psfrag{agq}[c][c][0.6]{90$^\circ$}
\psfrag{agh}[c][c][0.6]{60$^\circ$}
\psfrag{fa}[c][c][0.9]{(a)}
\psfrag{fb}[c][c][0.9]{(b)}
\psfrag{fc}[c][c][0.9]{(c)}
\psfrag{zb}[c][c][0.9]{$Z_\text{B}$}
\psfrag{yb}[c][c][0.9]{$Y_\text{B}$}
\psfrag{xb}[c][c][0.9]{$X_\text{B}$}
\psfrag{rqr1}[c][c][0.9]{$\bm{r}_1$~~~~~~~~~~~~~~~~~}
\psfrag{rqr2}[c][c][0.9]{$\bm{r}_2$}
\psfrag{rqr3}[c][c][0.9]{~~~~~~~$\bm{r}_3$}
\psfrag{rqr4}[c][c][0.9]{$\bm{r}_4$}
\psfrag{rtr1}[c][c][0.9]{$\bm{r}_1$}
\psfrag{rtr2}[c][c][0.9]{$\bm{r}_2$~~~~~~}
\psfrag{rtr3}[c][c][0.9]{$\bm{r}_3$~~~~~~~~~~~~~~~~~~}
\psfrag{rhr1}[c][c][0.9]{~~~~~~~~~~$\bm{r}_1$}
\psfrag{rhr2}[c][c][0.9]{$\bm{r}_2$}
\psfrag{rhr3}[c][c][0.9]{$\bm{r}_3$}
\psfrag{rhr4}[c][c][0.9]{$\bm{r}_4$~~~~~~}
\psfrag{rhr5}[c][c][0.9]{~~~~~~~~~~~~~~~~~~~~~~~~~$\bm{r}_5$}
\psfrag{rhr6}[c][c][0.9]{~~~~~~~~~~~~~~~~$\bm{r}_6$}
\psfrag{r5}[c][c][0.9]{$\bm{r}_5$}
\psfrag{r6}[c][c][0.9]{$\bm{r}_6$}
\psfrag{T1}[c][c][0.9]{$\bm{v}_1$}
\psfrag{T2}[c][c][0.9]{$\bm{v}_2$}
\psfrag{T3}[c][c][0.9]{$\bm{v}_3$}
\psfrag{T4}[c][c][0.9]{$\bm{v}_4$}
\psfrag{T5}[c][c][0.9]{$\bm{v}_5$}
\psfrag{T6}[c][c][0.9]{$\bm{v}_6$}
\includegraphics[width=160mm]{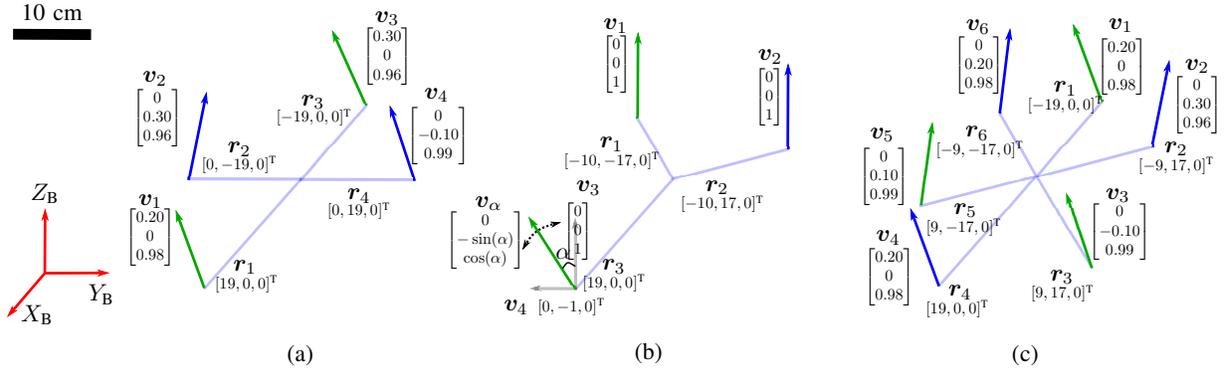}
\caption{The thrust orientations $\bm{v}_i$'s and the moment arms $\bm{r}_i$'s (in cm) of (a) quadrotor (b) tricopter and (c) hexacopter, where the green ones and blue ones denote the corresponding propellers rotating in clockwise (CW) direction and in the counterclockwise (CCW) direction, respectively.}\label{fig.tilt_thrusts}\vspace{-5mm}
\end{figure*}

\section{Numerical Verification}\label{sec:verification}

In this section, we show examples of the trajectory generation using the collocation method. Three multirotor robots with different configurations are explored.

\subsection{Trajectory Setpoints}
\begin{figure}[t]
\centering
\psfrag{ts}[c][c][0.9]{t (s)}
\psfrag{ag}[c][c][0.9]{$\bm{\theta}$ (rad)}
\psfrag{agt}[c][c][0.9]{angles (rad)}
\psfrag{phi}[c][c][0.9]{$\phi$}
\psfrag{the}[c][c][0.9]{$\theta$}
\psfrag{psi}[c][c][0.9]{$\psi$}
\psfrag{wx}[c][c][0.9]{$~\bm{\omega}_x$}
\psfrag{wy}[c][c][0.9]{$~\bm{\omega}_y$}
\psfrag{wz}[c][c][0.9]{$~\bm{\omega}_z$}
\psfrag{dag}[c][c][0.9]{$\bm{\omega}$ (rad/s)}
\psfrag{force}[c][c][0.9]{$\bm{u}_\text{n}$}
\psfrag{alpha}[c][c][0.9]{$\alpha$}
\psfrag{T1}[c][c][0.9]{$T_1$~~}
\psfrag{T2}[c][c][0.9]{~~$T_2$}
\psfrag{Ta}[c][c][0.9]{~~$T_\alpha$}
\subfigure[]{\includegraphics[width=42.6mm]{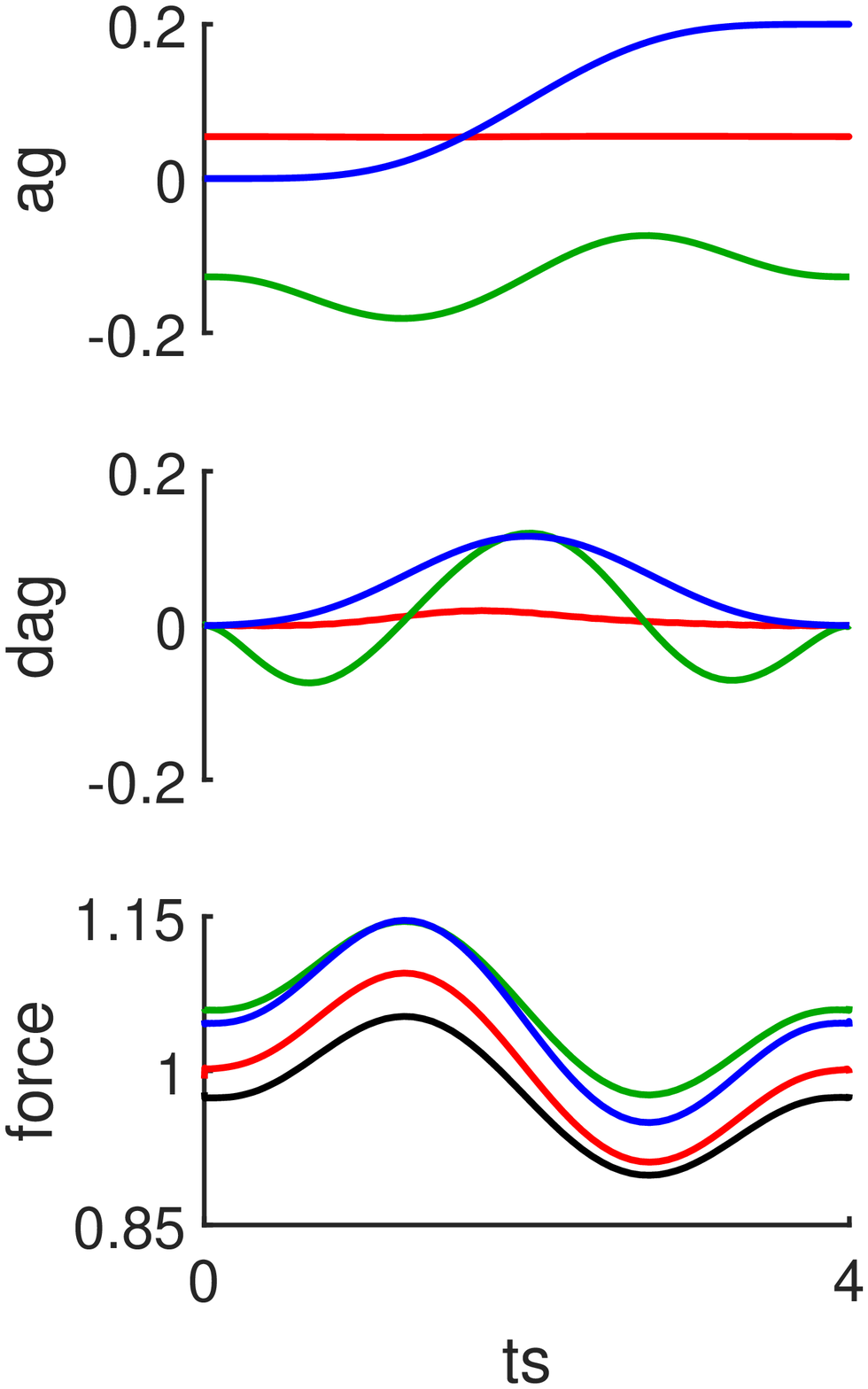}}
\subfigure[]{\includegraphics[width=42.6mm]{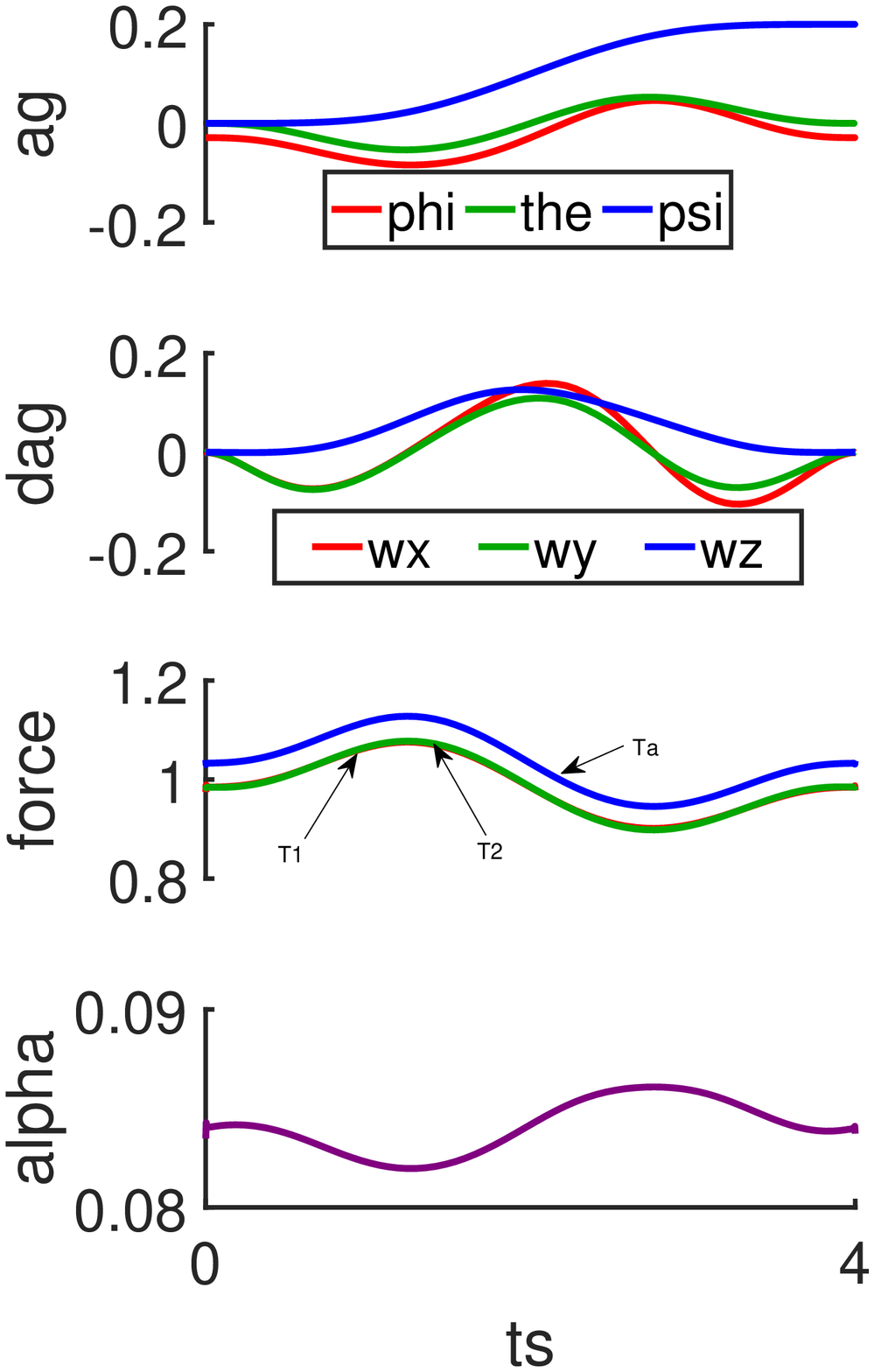}}
\caption{The states and control inputs of (a) quadrotor with $\text{rank}(\bm{A})=3$ and (b) conventional tricopter with $\text{rank}(\bm{A})=2$.}\label{fig.quadrotor_trajectory}\vspace{-4mm}
\end{figure}

The flight trajectory $\bm{\sigma}(t)$ (three positions and the heading angle) for all examples is generated for the time period $t\in[0, t_f=4\text{s}]$ using the polynomial path-primitives
\begin{equation}
\bm{\sigma}_i(t) = \sum^9_{j=0}d_{ij}t^j,\label{eqn.gnt_traj}
\end{equation}
where $d_{ij}$'s are the coefficients corresponding to an arbitrarily chosen trajectory: $\bm{\sigma}(0)=[0.0\text{m},0.0\text{m},0.0\text{m},0.0\text{rad}]^\text{T}$ to $\bm{\sigma}(t_f)=[-1.0\text{m},1.0\text{m},1.5\text{m},0.2\text{rad}]^\text{T}$; and the following boundary conditions for the time-derivatives of $\bm{\sigma}(t)$
\begin{equation}
\bm{\sigma}^{(i)}(t)=0~~\text{for }i=1,\cdots,4~~\text{when}~~t=0,t_{f}.\label{eqn.traj_cst}
\end{equation}
These conditions realistically restrict the aerial robots to start and end the trajectory in the hovering condition. The generated trajectory is illustrated in the Fig. \ref{fig.trajectory}.

\subsection{Example Vehicles with Different Rank Conditions}
\begin{figure}[ht]
\centering
\psfrag{ts}[c][c][0.9]{t (s)}
\psfrag{ag}[c][c][0.9]{$\bm{\theta}$ (rad)}
\psfrag{phi}[c][c][0.9]{$\phi$}
\psfrag{the}[c][c][0.9]{$\theta$}
\psfrag{psi}[c][c][0.9]{$\psi$}
\psfrag{wx}[c][c][0.9]{$~\bm{\omega}_x$}
\psfrag{wy}[c][c][0.9]{$~\bm{\omega}_y$}
\psfrag{wz}[c][c][0.9]{$~\bm{\omega}_z$}
\psfrag{dag}[c][c][0.9]{$\bm{\omega}$ (rad/s)}
\psfrag{force}[c][c][0.9]{$\bm{u}_\text{n}$}
\subfigure[]{\includegraphics[width=42mm]{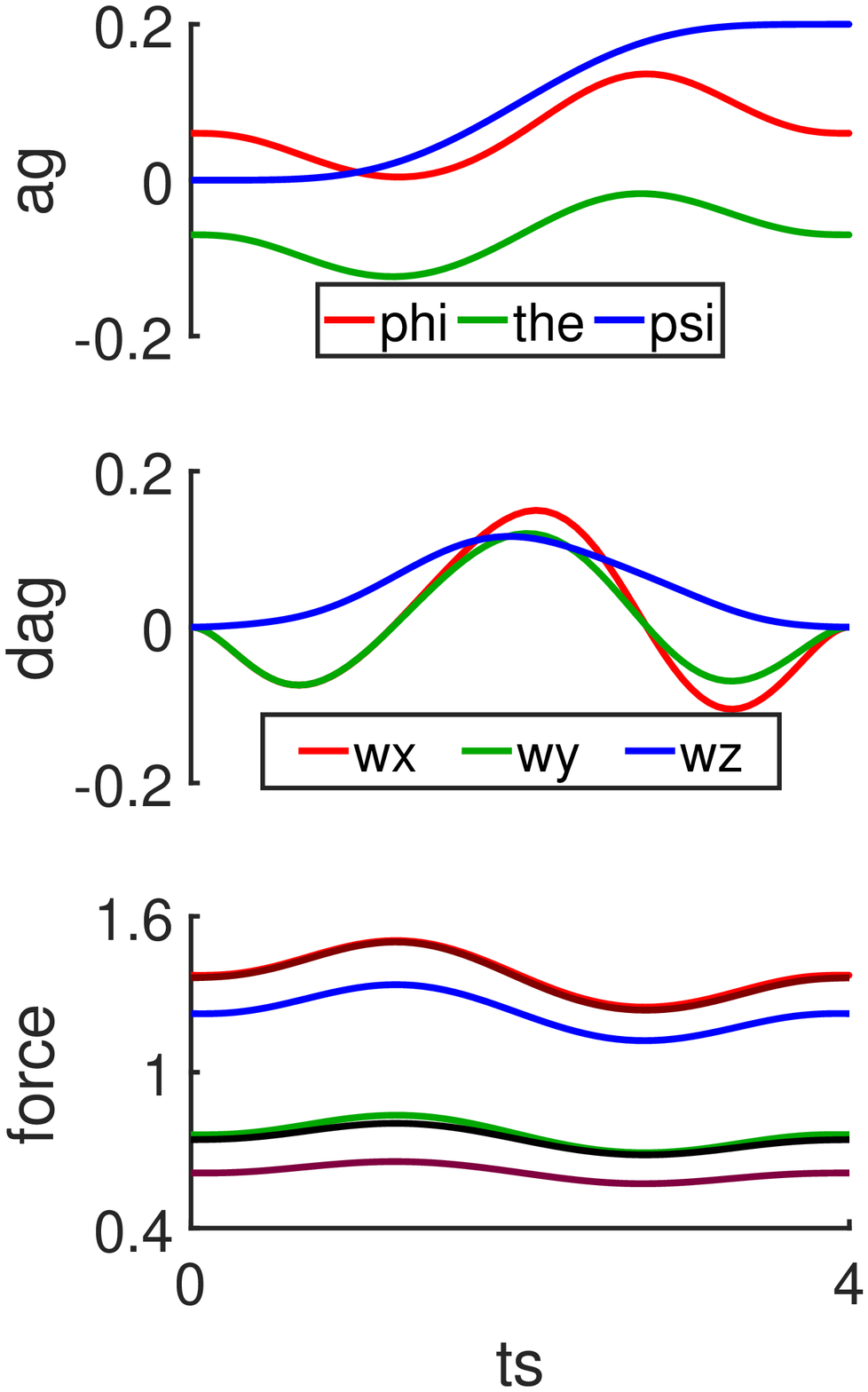}}
\subfigure[]{\includegraphics[width=42mm]{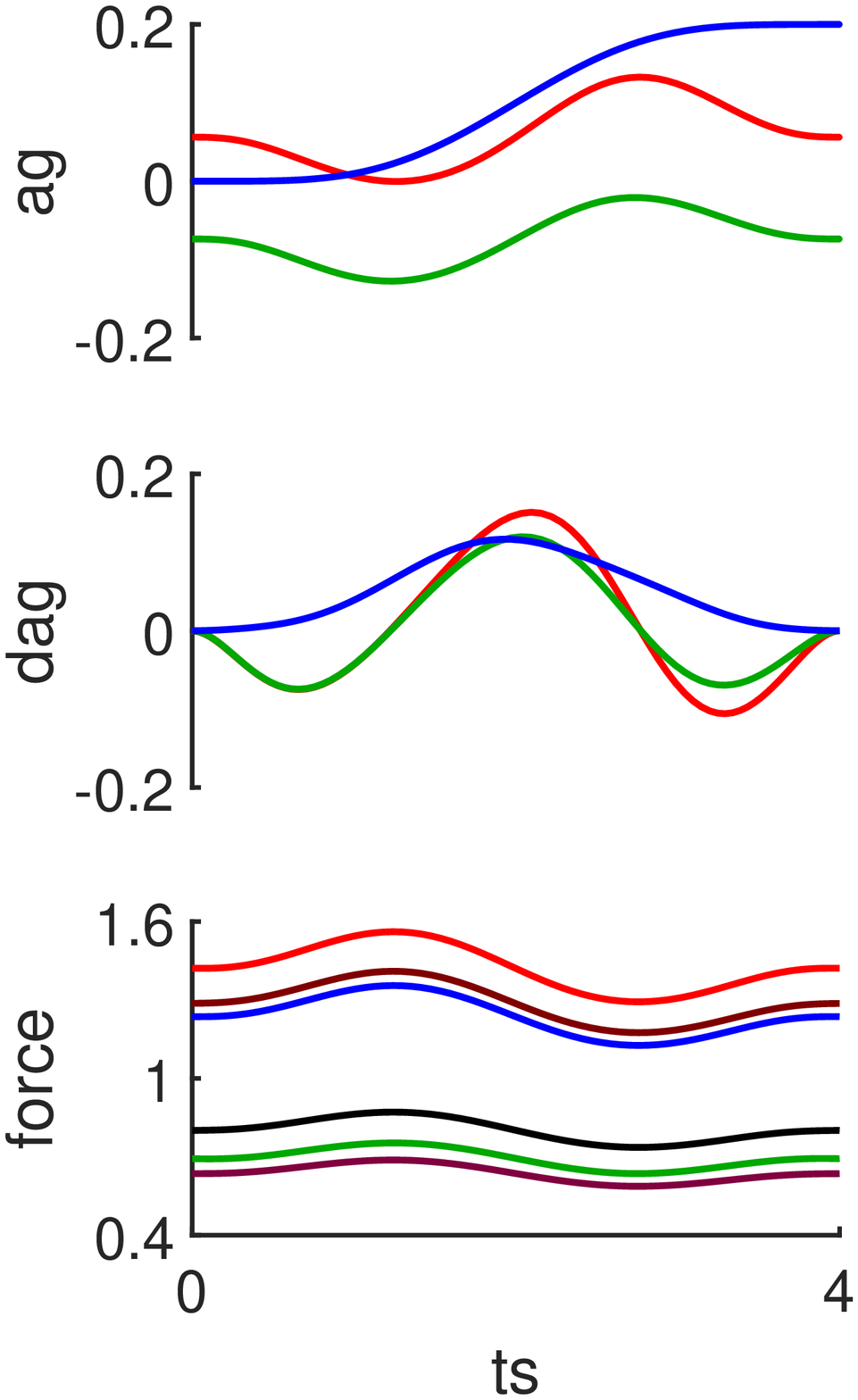}}
\caption{The states and control inputs of the hexacopter with tilted propellers by (a) adding extra outputs and (b) solving optimization problem.}\label{fig.hexacopter_t_trajectory}\vspace{-4mm}
\end{figure}

For demonstration, we select three representative aerial robots as shown in Fig. \ref{fig.tilt_thrusts}: a quadrotor, a tricopter, and a hexacopter. The quadrotor possesses four tilted propellers with $\text{rank}(\bm{A})=3$. The tricopter has three nominally propellers with one tilting arm. This renders the rank of $\bm{A}$ to be 2. For the hexacopter, $\text{rank}(\bm{A})=3$, but the number of propellers is six. This shows how our proposed method can be modified for cases where $N>4$.

For all robots, the basic configurations ($\bm{r}_i$'s and $\bm{v}_i$'s) are provided in Fig. \ref{fig.tilt_thrusts}. The masses are assumed to be $m=1.0$ kg, the inertias are $\bm{J}=\text{diag}(5,5,10)\times 10^{-3}$ kg$\cdot$m$^2$, and $c_i=\pm 1.6$ cm for all robots. For the ease of comparison, we define the normalized control input $\bm{u}_\text{n}:=N\bm{u}/{mg}$.
 
The nonlinear program for each robot is solved using a commercial solver (PROPT, Tomlab Optimization \cite{rutquist2010propt}) on Matlab (MathWorks). PROPT uses pseudospectral collocation methods and $100$ collocation points are applied.

\subsubsection{Quadrotor}

The thrust vectors of the quadrotor in Fig. \ref{fig.tilt_thrusts}(a) are all unparalleled, resulting in a full-rank $\bm{A}$. As stated in Section \ref{sec.trajectory_gen}, the direction of the total thrust varies with respect to the body frame as the total torque changes.

With the chosen trajectory, the outcomes of the trajectory generation are shown in Fig. \ref{fig.quadrotor_trajectory}(a). As anticipated, the quadrotor starts and ends its trajectory in the hovering condition, with non-zero Euler angles. The state variables (except the yaw angle $\bm{\sigma}_4$) are identical at the beginning and the end of the trajectory. The $2^\text{nd}$ and $3^\text{rd}$ propellers are required to generate larger thrusts than the $1^\text{st}$ and $4^\text{th}$ propellers due to their larger tilt angles with respect to $Z_B$. On average, combined thrust is larger than the robot's weight.

\subsubsection{Tricopter}

The tricopter has three propellers which are nominally vertically directed. As shown in Fig. \ref{fig.tilt_thrusts}(b), the third arm is actuated. Its rotation is described by the angle $\alpha$ such that $\bm{v}_\alpha=[0,-\sin\alpha,\cos\alpha]^\text{T}$ and $\text{rank}(\bm{A})=2$.

With three thrust commands ($T_1$, $T_2$, and $T_\alpha$) and one rotational input $\alpha$, the dynamics of the system with four inputs can be transformed to fit into our framework by defining two hypothetical orthogonal thrust directions: $\bm{v}_3=[0,0,1]^\text{T}$ and $\bm{v}_4=[0,-1,0]^\text{T}$ (see Fig. \ref{fig.tilt_thrusts}(b)). The thrust generated by the tilting arm, $\bm{v}_\alpha T_\alpha$, becomes $\bm{v}_3T_3+\bm{v}_4T_4$ for the newly defined inputs $T_3$ and $T_4$. The tricopter, consequently, can be mathematically treated as a quadrotor with $\bm{r}_3=\bm{r}_4$.

The outcomes of trajectory generation are illustrated in Fig. \ref{fig.quadrotor_trajectory}(b). The tricopter starts and ends its trajectory in the hovering condition, with non-zero Euler angles. The plot reveals that $T_\alpha > T_1,T_2$ and $T_1\approx T_2$ for the whole flight. It can be deduced that the yaw torques generated by the $1^{\text{st}}$ and $2^{\text{nd}}$ propellers more or less cancel out. Accordingly, the total yaw torque contributed by $T_\alpha$ is approximately zero as the aerodynamic drag torque is balanced out by the tilted arm. This is consistent with the fact that the tilt angle $\alpha$ always changes in the opposite direction to its the normalized thrust $\bm{u}_{\text{n},\alpha}$.

\subsubsection{Hexacopter}

Despite having a full-rank $\bm{A}$, unlike the previous two vehicles, the hexacopter in Fig.\ref{fig.tilt_thrusts}(c) has six propellers ($N=6$). Some adjustment is necessary to adopt our proposed trajectory generation strategy. We present two approaches in the followings.

The first approach is to define two additional (flat) outputs so that the number of the constraints are equal to the number of control inputs. We opt to define ${\sigma}_5$ and ${\sigma}_6$ such that the resultant dynamics of the hexacopter resemble that of a conventional quadrotor. This is achieved by ensuring the direction of the total thrust $\bm{A}\bm{u}$ remains unchanged during the whole flight. Mathematically, this translates to
\begin{equation}
\begin{split}
{\sigma}_5=[1,0,0]\frac{\bm{A}\bm{u}}{\|\bm{A}\bm{u}\|},~~~~
{\sigma}_6=[0,1,0]\frac{\bm{A}\bm{u}}{\|\bm{A}\bm{u}\|}, \label{eqn.extra_sigma}
\end{split}
\end{equation}
where we choose $\sigma_5=0.07$ and $\sigma_6=0.06$. In this particular case, the outputs are functions of both $\bm{x}$ and $\bm{u}$. The caveat of this method is that, with the extra outputs, the proof for the existence of the solution in Section \ref{sec.existence} no longer applies. If these additional outputs are not true flat outputs of the system, a feasible solution does not exist. It requires some knowledge from the designer to define reasonable outputs. 

The second method is to exploit the two unconstrained DoFs of the control inputs and turn the trajectory generation problem into a trajectory optimization problem. This way, a feasible trajectory is still guaranteed to exist. Moreover, the implementation is relatively simple as conventional software packages for solving collocation problems are designed for trajectory optimization. In this case, we set up the optimization problem using the following quadratic input cost
\begin{figure}[t]
\centering
\psfrag{ts}[c][c][0.9]{t (s)}
\psfrag{sgm5}[c][c][0.9]{$\sigma_5$}
\psfrag{sgm6}[c][c][0.9]{$\sigma_6$}
\psfrag{sg56}[c][c][0.9]{$\sigma_5$,~~$\sigma_6$}
\psfrag{sg56a}[c][c][0.9]{(a)}
\psfrag{sg56b}[c][c][0.9]{(b)}
\subfigure[]{\includegraphics[width=40mm]{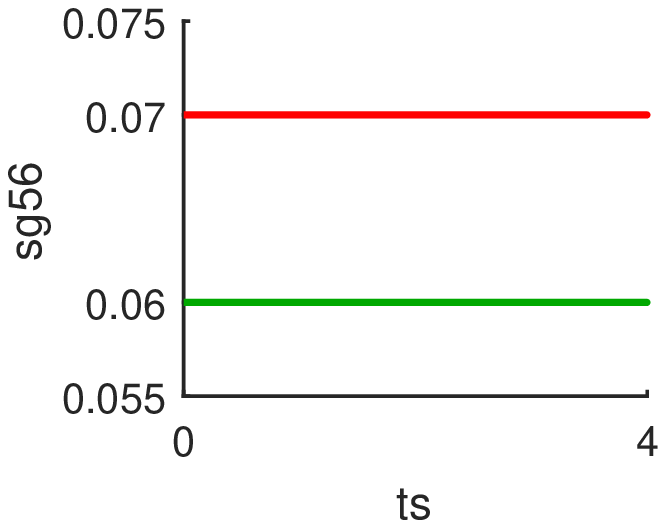}}
\subfigure[]{\includegraphics[width=40mm]{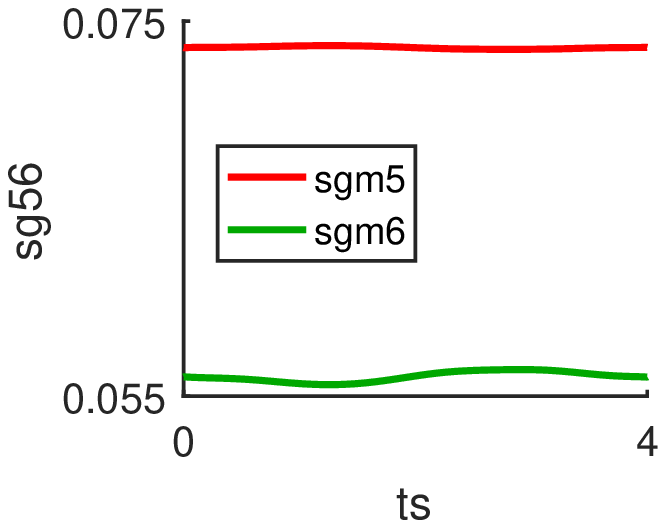}}
\caption{Comparison of the $\sigma_5$ and $\sigma_6$ for the hexacopter in two methods: (a) adding extra outputs and (b) solving an optimization problem.}\label{fig.sg56}
\vspace{-4mm}
\end{figure}
\begin{equation}
\min_{\bm{u}_\text{n}}~~~~\frac{1}{N t_f}\int^{t_f}_{t=0}\bm{u}_\text{n}^\text{T}\bm{u}_\text{n} \rm{d}t, \label{eqn.objectivefun}
\end{equation}
subject to the system dynamics and the desired flat outputs. For a conventional quadrotor in a perfect hovering condition, the expected cost is unity.

The generated states and control inputs for the hexacopter from both strategies are shown in Fig. \ref{fig.hexacopter_t_trajectory}. Due to the asymmetric configuration, some propellers are required to generate larger thrusts than others. At the first glance, we observe minor differences between the results. A closer inspection into Fig. \ref{fig.sg56}(a) verifies that, for the first method, where we specifically defined $\sigma_5$ and $\sigma_6$, their values remain constant throughout the flight as commanded. As anticipated, during the process, we find that the solutions do not always exist if the values of $\sigma_5$ and $\sigma_6$ are poorly chosen. For comparison, we plot the corresponding values of $\sigma_5$ and $\sigma_6$ of the trajectory generated by minimizing the cost function in Fig. \ref{fig.sg56}(b). These numbers show a slight variation over time, indicating a modest change in the direction of the thrust vector in the body frame.

The quadratic costs of two trajectories in Fig.\ref{fig.hexacopter_t_trajectory} as defined by equation \eqref{eqn.objectivefun} are $1.130$ and $1.127$. The optimization reduces the cost by $\approx3\%$. While this number is insignificant and unlikely to matter in practice, the primary advantage of the second method is the guaranteed existence of a feasible solution.

\section{Conclusion and Discussion}

We have investigated the dynamics of multirotor vehicles with tilted propellers based on the differential flatness theory and shown that the position and heading angle are the flat outputs. Similar to conventional quadrotors, the flatness property has a radical implication on the trajectory generation and controllability. The result essentially means that this class of underactuated vehicles is capable of following an arbitrary trajectory in 3D space. In addition, we have proposed a universal method of trajectory generation for computing the corresponding states and control inputs for given flat outputs. With the aid of software package for trajectory collocation, we verify the proposed strategy using three different robots, including a quadrotor, a tricopter and a hexacopter with unparalleled thrusts. In practice, the feasibility of a trajectory should be considered with other performance requirements and physical constraints \cite{brescianini2018computationally,morrell2018differential}. 

The outcomes of this paper pave the way for future work on flight control. Despite the existence of feasible trajectories, novel controllers must be develop to accommodate the dynamics of vehicles with unaligned thrusts, which are notably different from those of traditional multirotor robots. In contrast to previous works on this class of robots that only focus on attitude stabilization or rely on linearization near the hovering state \cite{efraim2015quadrotor,ansari2017adaptive}, this composes a nontrivial task that must be accomplished for a trajectory following flight demonstration.
%\section*{Acknowledgment}
%
%This work is supported by ...



%\bibliographystyle{IEEEtran}
\vspace{-2mm}
\bibliographystyle{IEEEtran}
\bibliography{references}

\end{document}